\documentclass[journal]{IEEEtran}
\usepackage{graphicx}
\usepackage{booktabs}
\usepackage{times}
\usepackage{soul}
\usepackage{color}
\usepackage{xcolor}
\usepackage{url}
\usepackage{amsfonts,amssymb}
\usepackage[hidelinks]{hyperref}
\usepackage[utf8]{inputenc}
\usepackage[small]{caption}
\usepackage{graphicx}
\usepackage{amsmath}
\usepackage{amsthm}
\usepackage{booktabs}
\usepackage{algorithm}
\usepackage{xcolor} 
\usepackage{algorithmic}
\usepackage[switch]{lineno}
\usepackage{booktabs}
 \usepackage{multirow}
 \usepackage{pifont}
 \usepackage{bm}
 \usepackage{tabularx}
 \usepackage{adjustbox}
 \usepackage{makecell}
\usepackage{array}  
\usepackage{caption}
\usepackage{multirow}
\usepackage{colortbl}

\usepackage{tabularx}  
\usepackage{xcolor}    

\ifCLASSINFOpdf
\else
\fi

\def\ourmethod{GLFM}

\begin{document}

\title{Boosting Global-Local Feature Matching via Anomaly Synthesis for Multi-Class Point Cloud Anomaly Detection}

\author{Yuqi~Cheng,~\IEEEmembership{Student Member, IEEE},
        Yunkang~Cao,~\IEEEmembership{Graduate Student Member, IEEE},
        Dongfang~Wang,
        
        Weiming Shen\IEEEauthorrefmark{1},~\IEEEmembership{Fellow,~IEEE},
        Wenlong~Li,~\IEEEmembership{Member,~IEEE}
\thanks{Weiming Shen\IEEEauthorrefmark{1} (wshen@ieee.org) is the corresponding author.}
\thanks{
This work was supported by Fundamental Research Funds for the Central Universities (HUST: 2021GCRC058) and was part by the HPC Platform of Huazhong University of Science and Technology where the computation is completed.
Yuqi Cheng, Yunkang Cao, Dongfang~Wang, Weiming Shen, and Wenlong Li are with the State Key Laboratory of Intelligent Manufacturing Equipment and Technology, Huazhong University of Science and Technology, Wuhan 430074, China.(e-mail: yuqicheng@hust.edu.cn; cyk\_hust@hust.edu.cn; wdf@hust.edu.cn; wshen@ieee.org; wlli@mail.hust.edu.cn).

}}

\markboth{Finished in May 2024, Accepted by IEEE TASE in February 2025}%
{Shell \MakeLowercase{\textit{et al.}}:  Bare Demo of IEEEtran.cls for IEEE Journals}
\maketitle

\IEEEpeerreviewmaketitle
\begin{abstract}

Point cloud anomaly detection is essential for various industrial applications. The huge computation and storage costs caused by the increasing product classes limit the application of single-class unsupervised methods, necessitating the development of multi-class unsupervised methods. However, the feature similarity between normal and anomalous points from different class data leads to the feature confusion problem, which greatly hinders the performance of multi-class methods. Therefore, we introduce a multi-class point cloud anomaly detection method, named GLFM, leveraging global-local feature matching to progressively separate data that are prone to confusion across multiple classes. Specifically, GLFM is structured into three stages: Stage-I proposes an anomaly synthesis pipeline that stretches point clouds to create abundant anomaly data that are utilized to adapt the point cloud feature extractor for better feature representation. Stage-II establishes the global and local memory banks according to the global and local feature distributions of all the training data, weakening the impact of feature confusion on the establishment of the memory bank. Stage-III implements anomaly detection of test data leveraging its feature distance from global and local memory banks. Extensive experiments on the MVTec 3D-AD, Real3D-AD and actual industry parts dataset showcase our proposed GLFM's superior point cloud anomaly detection performance. The code is available at \href{https://github.com/hustCYQ/GLFM-Multi-class-3DAD}{\url{https://github.com/hustCYQ/GLFM-Multi-class-3DAD}}.
\end{abstract}





\section{Introduction}

\begin{figure}[t!]
\centering\includegraphics[width=\linewidth]{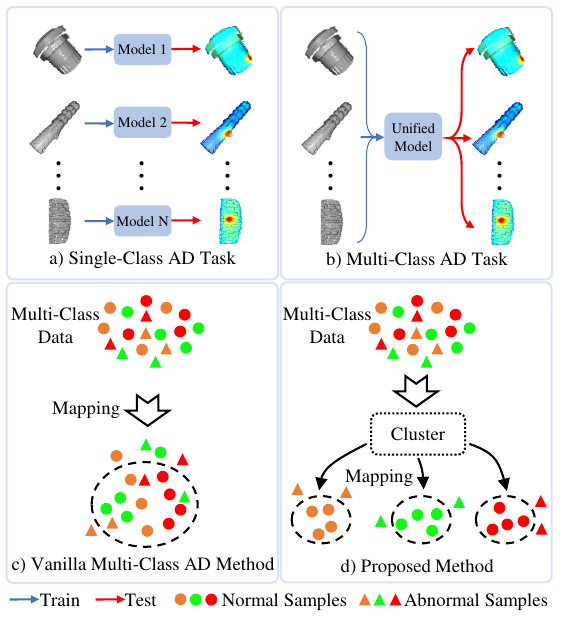}
\caption{a) Illustration of the single-class AD task, in which each class needs a specialized model. b) Illustration of the multi-class AD task, in which a unified model is trained and tested in the data from multiple classes. c) Vanilla multi-class AD method struggles to distinguish normal and anomaly data due to feature confusion. d) The proposed method clusters by global features and performs AD on each class to solve feature confusion.}\vspace{-5mm}
\label{fig:framework}
\end{figure}

\IEEEPARstart{P}{oint} cloud anomaly detection (AD) is crucial for identifying geometric defects in industrial parts~\cite{xie2023iad,MVGR}. Owing to the challenges in collecting anomaly samples compared to normal samples, unsupervised anomaly detection methods~\cite{liu2024deep,KD_Luo}, which necessitate only normal data for training, have become the predominant solution. Current point cloud AD methods adhere to this unsupervised paradigm, training specialized detection models for each class, as illustrated in Fig.\ref{fig:framework} a). However, the necessity to inspect a multitude of products, coupled with the rising number of classes, leads to substantial costs associated with the development of specialized models~\cite{Diad,patchcore}. To enhance the deployment ability of AD methods, recent advancements have adopted a multi-class AD scheme~\cite{MVP-PCLIP}. This approach involves training a model on data from multiple classes and detecting anomalies across these classes, as depicted in Fig.~\ref{fig:framework} b).

\begin{figure}[t!]
\centering\includegraphics[width=0.8\linewidth]{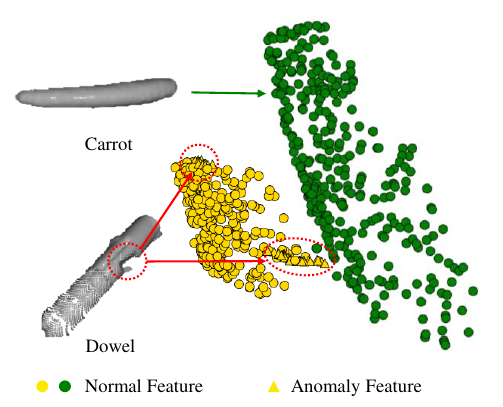}
\caption{\textbf{Feature visualization of anomaly data from Dowel class and normal data from Carrot class.} Features are extracted through PointMAE~\cite{pointmae} that is pre-trained on ShapeNet~\cite{Shapenet} and principal components analysis (PCA) is used to reduce the feature dimension to two for visualization. Yellow shapes represent the features of Dowel data, and green shapes represent the features of Carrot. Anomalous regions/features are highlighted within a red box.}
\vspace{-5mm}
\label{fig:pca}
\end{figure}

However, certain normal patterns in one class may appear as anomalies in another, complicating accurate anomaly detection. For example, as presented in Fig.~\ref{fig:pca}, while anomalous features of Dowel data deviate from the normal features, some of them can closely resemble the normal features of the Carrot class. Consequently, certain anomalous points may be mistakenly classified as normal during detection, which is referred to as feature confusion between different classes. This limitation hinders the distinction between normal and anomalous features based solely on local features or simply combining global and local features, thus restricting the performance of multi-class point cloud AD. Existing multi-class AD methods~\cite{unionAD,HVQ,ViTAD} struggle to address the feature confusion problem due to limited feature representation, as depicted in Fig.~\ref{fig:framework} c).

Global features, which encompass larger receptive fields, provide more descriptive information and aid in diminishing inter-class feature confusion. Conversely, local features, being more sensitive to anomalies, enable the detection of anomalies both object-wise and point-wise. Hence, we propose a Global-Local Feature Matching (\ourmethod{}) strategy to address the multi-class feature confusion problem in multi-class AD tasks. This strategy leverages global features from the training data to perform clustering into distinct classes, subsequently employing local features for unsupervised anomaly detection within each class. Enhancing feature descriptiveness is also crucial for addressing feature confusion. Specifically, we propose an anomaly synthesis pipeline to improve the feature descriptiveness of feature extractors through self-supervised learning, thereby rendering normal and abnormal features more discriminable. Experimental results on MVTec 3D-AD, Real3D-AD, and real-world applications demonstrate the superiority of our proposed method.

In summary, the contributions of this study are summarized as follows:

\begin{itemize}

    \item To the best of our knowledge, we are the first to explore multi-class point cloud anomaly detection, aiming to construct a unified anomaly detection model for multi-class point clouds. 
    \item We identify feature confusion in the multi-class AD task and propose a Global-Local Feature Matching (\ourmethod{}) strategy for multi-class point cloud anomaly detection, sequentially utilizing global and local features to effectively address the problem of feature confusion.
    \item We propose an anomaly data synthesis pipeline to generate realistic 3D anomalies, which can be utilized to adapt the feature extractor by self-supervised learning and improve the feature expression ability of anomaly regions.

\end{itemize}

The rest of this paper is organized as follows. Section~\ref{sec:related-work} comprehensively reviews the related work. The framework and key technologies of the proposed method \ourmethod{}, are elaborated in Section~\ref{sec:method}. In Section~\ref{sec:exp}, the experiments are carried out to evaluate the performance of the proposed method. The conclusion is provided in Section~\ref{sec:conclusion}.

\section{Related Work}\label{sec:related-work}
In this section, we review topics closely related to the technical aspects of this paper, including single-class anomaly detection, multi-class anomaly detection, and self-supervised learning.

\subsection{Single-Class Anomaly Detection}

The investigation of unsupervised AD methods begins in the image field~\cite{RAD,10884560}, which can be divided into three categories, including flow-based methods~\cite{rudolph2021same,FL_Yao}, knowledge-distillation-based methods~\cite{VarAD,KD_Luo2} and memory-bank-based methods~\cite{Liu_MB,MB_yao,GCPF}. Following image AD, point cloud AD methods also develop knowledge-distillation-based methods and memory-bank-based methods. Knowledge-distillation-based methods~\cite{Autoencoder,ST3D} build a teacher-student framework, utilizing the inconsistent features between features extracted by the teacher model and student model to detect anomalies. In contrast, 3D-ST~\cite{STdes} performs self-supervised learning on the teacher network, enabling it to better describe point cloud features. However, the overall performance of knowledge-distillation-based methods still has significant shortcomings. Memory-bank-based methods~\cite{BTF,M3DM,cpmf,shape} store the extracted normal data features in a bank and determine anomalies by computing the distance between the extracted features and the features of the bank during testing. The difference between these methods lies in the way they extract features: BTF~\cite{BTF} uses handcrafted FPFH (Fast Point Feature Histograms) features; M3DM~\cite{M3DM} utilizes pre-trained PointMAE~\cite{pointmae} model to extract features; CPMF~\cite{cpmf} projects point clouds into multi-view images and extracts features using a pre-trained image encoder; Shape-guided~\cite{shape} considers PointNet \cite{PointNet} and NIF \cite{NIF} to learn local representation of surface geometry. Currently, memory-bank-based methods achieve superior point cloud anomaly detection performance in comparison to other schemes like knowledge-distillation-based methods.

However, the above-mentioned methods are typically developed for single-class AD, and face challenges when applied to multi-class AD due to neglecting the problem of feature confusion between different classes.

\subsection{Multi-Class Anomaly Detection}

Multi-class image AD has recently received a lot of attention. UniAD~\cite{unionAD} reconstructs anomalous features as normal features and utilizes the differences in features between before and after reconstruction to detect anomalies. It identifies the challenge of multi-class AD as “identical short-cut”. To further prevent “identical short-cut”, OmniAL~\cite{OmniAL} introduces synthetic anomalies, enhancing the decoder's ability to reconstruct anomalous features into normal features; HVQ~\cite{HVQ} proposes a vector quantize based transformer and induces large feature discrepancy for anomalies. Instead, MSTAD~\cite{MSTAD} strives to maintain distinct common embeddings for individual categories and leverages the embeddings to revert anomalous features to normal features. Besides, PSA-VT~\cite{MC_Yao} reconstructs multi-scale discriminated local representation through proposed long-range global semantic aggregation. In addition, a more powerful reconstruction model, the diffusion model, has also been used in multi-class AD~\cite{Removing,Diad}. These models introduce random noise into normal data and train autoencoders as denoising models. Anomalies can be considered as noise to be removed and normal features are ultimately reconstructed. Furthermore, large visual models such as ViT~\cite{ViT} and vision-language models such as CLIP~\cite{CLIP} have also been used for anomaly reconstruction.

The above methods aim to distinguish features of normal and anomaly data from multiple classes. However, due to the feature confusion mentioned earlier, normal and anomaly data are not easily distinguishable in the feature space. Although SelFormaly~\cite{SelFormaly} attempts to mitigate feature confusion by utilizing the top $K$ features instead of the top one feature to compute anomalies, it still results in significant errors when the feature bank contains many similar features. Moreover, unlike images containing rich texture information, point clouds are susceptible to local geometric similarities, leading to more severe feature confusion. To address this, we propose a global-local feature matching strategy that progressively utilizes global and local features to distinguish locally similar data, weaken the feature confusion, and enhance the performance of multi-class point cloud anomaly detection.

\subsection{Self-supervised Learning}
Self-supervised learning, which uses unlabeled data to construct proxy tasks, can help models learn useful features for downstream tasks. In image anomaly detection, the discriminate ability of the model can be improved by synthesizing abnormal data and constructing self-supervision. Cutpaste~\cite{Cutpaste} creates local image inconsistencies considered as anomalies by cutting patches from a normal image and pasting them onto other regions. Synthesizing more realistic anomaly data benefits the training of better feature extractors. Subsequent works~\cite{Natural,Draem,realnet,DFMGAN} have aimed to synthesize more realistic anomalies. NSA~\cite{Natural} uses Poisson image editing to create more natural anomaly regions. DRAEM~\cite{Draem} leverages a public texture dataset to synthesize various texture anomalies. RealNet~\cite{realnet} proposes a diffusion-based synthesis strategy to generate data with varying anomaly strengths, simulating real anomaly distributions. Additionally, DFMGAN~\cite{DFMGAN} utilizes a few real anomalies to produce more realistic synthetic anomaly data.

However, due to the discrete and irregular nature of point clouds, the above methods are difficult to migrate to the point cloud data. This makes synthesizing locally continuous and smooth anomaly data particularly challenging. Although Anomaly-ShapeNet~\cite{shape_anomaly} acquires point cloud anomaly data by editing the ShapeNet~\cite{Shapenet} dataset, such editing cannot be applied to discrete points and can only be done on 3D models that are often not available. Therefore, to obtain a substantial and diverse amount of anomaly data for the point cloud feature extractor's self-supervised learning, we propose an automatic anomaly synthesis pipeline that stretches along the normal direction at any position of point clouds to create protrusion or depression defect.

\section{\ourmethod{} Method}\label{sec:method}
\subsection{Problem Definition}

Multi-class point cloud AD aims to develop a unified model capable of identifying anomalies across various classes. In this context, normal data from multiple classes are employed for training. During testing, an object-wise anomaly score $\xi \in [0,1]$ and a point-wise anomaly map $\boldsymbol{A} \in \mathbf{R}^{n \times 1}$ are computed for point clouds $\boldsymbol{P} \in \mathbf{R}^{n \times 3}$ from the training classes, where $n$ denotes the number of points. Higher values of $\xi$ and $\boldsymbol{A}$ signify higher anomaly levels.

\subsection{Overview}

\begin{figure*}[t!]
\centering\includegraphics[width=\linewidth]{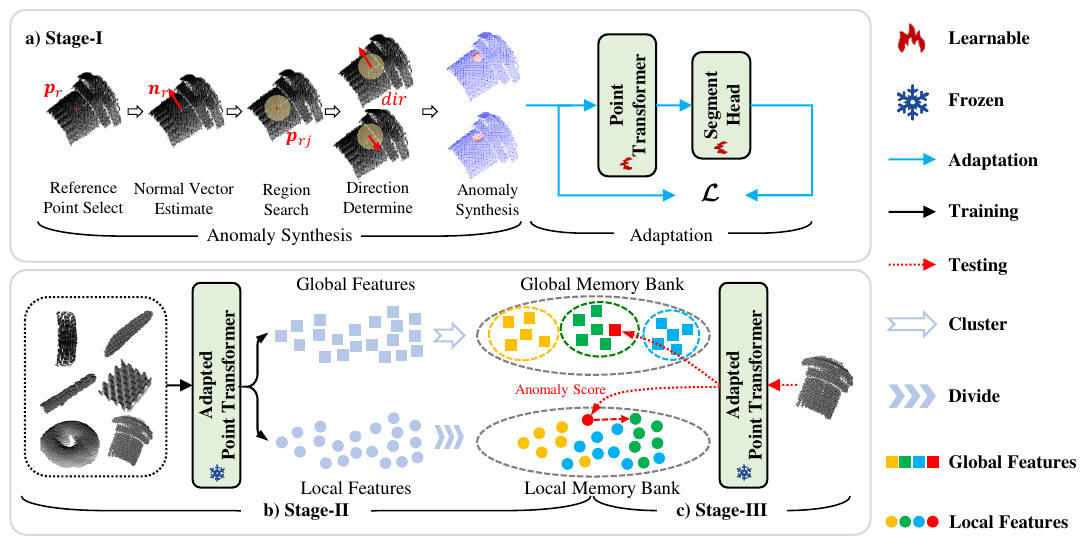}
\caption{\textbf{The framework of the proposed \ourmethod{}.} a) Stage-I: Anomaly data and corresponding point-wise labels are synthesized by select part points of the point clouds for irregular stretching to produce local protrusions or depressions. Then point cloud feature extractor (point transformer) is adapted by self-supervised learning of point cloud anomaly segmentation. b) Stage-II: During the training process, the global and local features of all training data are extracted by point cloud feature extractor adapted. Global features are clustered to construct $\boldsymbol{\mathcal{M}}_{G}$. Local features of training data are divided into multiple local memory banks $\boldsymbol{\mathcal{M}}_{i}$ based on the distance between the cluster centers and their global features. c) Stage-III: During the testing process, both global and local features of input point cloud are extracted seems to training process, and $\boldsymbol{\mathcal{M}}_{G}$ is used to answer the query of global features. Local features are then used to perform anomaly detection in the corresponding $\boldsymbol{\mathcal{M}}_{i}$.
}
\label{fig:method_overview}
\end{figure*}

The proposed method \ourmethod{} adapts a pre-trained point cloud feature extractor to extract global and local features from the data, and the anomaly detection of multi-class point cloud data is achieved by sequentially established global and local memory banks. \ourmethod{} consists of three stages, as shown in Fig.~\ref{fig:method_overview}. An anomaly synthesis pipeline is proposed in Stage-I to generate synthetic anomaly data, facilitating the construction of a generalized point cloud feature extractor through self-supervised learning. Stage-II establishes the global and local memory banks with the extracted global and local features, addressing the feature confusion caused by the similarity between normal and anomalous features from different classes. Stage-III detects anomalies through the comparison of test data features against those stored in the global memory bank and the corresponding local memory banks.

\subsection{Stage-I Pre-trained Backbone Adaptation}
Memory-bank-based methods typically employ a pre-trained feature extractor to extract the features of training data to build a memory bank. However, existing pre-trained point cloud backbones are trained on general datasets that have significantly different distributions to our targeted industrial anomaly detection data, leading to the neglect of industrial anomalous feature extraction. To improve feature descriptiveness, we propose to synthesize realistic anomalies to adapt the pre-trained backbones.

\subsubsection{Anomaly data synthesis}

Real-world 3D anomalies are typically local variations, so we consider incorporating local protrusions or depressions as geometric deformation defects in the point cloud, as shown in Fig.~\ref{fig:method_overview} a). A reference point is selected and its normal vector is estimated. Then the neighboring points of the reference point are stretched to generate protrusions or depressions according to different normal vector directions. The specific steps are as follows.

Assuming the input normal point cloud $\boldsymbol{P}\in \mathbf{R}^{n\times3}$ has $n$ points and $\boldsymbol{p}_r$ is the reference point selected from $\boldsymbol{P}$. The nearest neighbor search method (KNN) is utilized to find the top $k$ nearest neighbor points $\boldsymbol{p}_{rj}, 1 \leq j \leq k$ of point $\boldsymbol{p}_r$:

\begin{equation}
[\boldsymbol{p}_{r1},\boldsymbol{p}_{r2},\cdots, \boldsymbol{p}_{rk}]=KNN(\boldsymbol{p}_{r})\label{F1}
\end{equation}
Consider the top $k$ nearest neighbor points as a subset $\boldsymbol{Q}_r$, and use principal component analysis (PCA) to calculate the eigenvectors of $\boldsymbol{Q}_r$:
\begin{equation}
[\boldsymbol{V}_{1},\boldsymbol{V}_{2},\boldsymbol{V}_{3}]=PCA(\boldsymbol{Q}_{r})\label{F2}
\end{equation}
where $\boldsymbol{V}_{1},\boldsymbol{V}_{2},\boldsymbol{V}_{3}$ are the eigenvectors of $\boldsymbol{Q}_{r}$. $\lambda_1, \lambda_2, \lambda_3$ are the eigenvalues corresponding to the three eigenvectors and satisfying $\lambda_1> \lambda_2 > \lambda_3$. Therefore, the eigenvector $\boldsymbol{V}_{3}$ that corresponds to the minimum eigenvalue $\lambda_3$ is the normal vector $\boldsymbol{n}_r$ of the point $\boldsymbol{p}_r$~\cite{cheng2022novel}.

Next, we search the top $C$ nearest neighbor points $\boldsymbol{p}_{rj}, 1 \leq j \leq C$ and stretch them as anomalies.
To increase the size diversity of stretched points, $C$ is set as a random variable related to the total point number of the point cloud and follows a uniform distribution. The stretched distance is determined by the local density $\rho_r$ of the point cloud. $\rho_r$ is calculated by:
\begin{equation}
\rho_r=||\boldsymbol{p}_{r1}-\boldsymbol{p}_{r2}||_2\label{F3}
\end{equation}
where $\boldsymbol{p}_{r1}$ and $\boldsymbol{p}_{r2}$ are the closest point and second closest point to $\boldsymbol{p}_{r}$, respectively. The total stretched distance $D_r$ is calculated by:

\begin{equation}
D_r=\rho_r \cdot C\label{F4}
\end{equation}
The offset of each point is determined by the distance from point $\boldsymbol{p}_{r}$, and then the points stretched $\boldsymbol{p}_{rj}^{'}$ are:

\begin{equation}
\boldsymbol{p}_{rj}^{'}=\boldsymbol{p}_{rj} + dir \cdot \boldsymbol{n}_r \cdot D_r \cdot \frac{j}{C} \label{F5}
\end{equation}
where $dir$ is the direction of stretching, $dir=1$ means generating protrusion and $dir=-1$ means generating depression.

In addition, to enhance the shape diversity of stretched points, three weights are added to the coordinate system units in search of $\boldsymbol{p}_{rj}$:

\begin{equation}
\left\{
\begin{aligned}
X^{'} &= X \cdot dx  \\
Y^{'} &= Y \cdot dy\\
Z^{'} &= Z \cdot dz\\
\end{aligned}
\right.\label{F6}
\end{equation}
where $X,Y,Z$ are the coordinate system units of the original point cloud, $X^{'}, Y^{'}, Z^{'}$ are the coordinate system units added weights, $dx, dy, dz$ are the weights for the three axes, respectively, following a uniform distribution.

\begin{figure}[h!]
\centering\includegraphics[width=\linewidth]{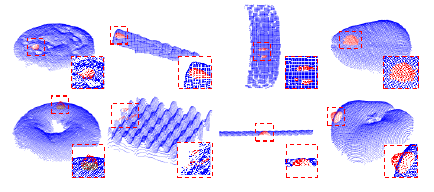}
\caption{\textbf{Visualization of synthetic anomaly data from different classes.} The red points represent the anomalies. The synthetic anomaly data have the properties of local continuity and smoothness.}
\label{fig:anomaly}
\end{figure}

The synthetic anomaly data $\boldsymbol{P}^{'}$ are shown in Fig.~\ref{fig:anomaly}. Although the anomaly data exhibit local protrusions or depressions through stretching, the overall distribution of the point cloud still remains coherent, thereby ensuring similarity with the actual point cloud. The overall process is summarized in Algorithm \ref{alg:gen}.

\begin{algorithm}[!h]
    \caption{Algorithm of Anomaly Synthesis}
    \label{alg:gen}
    \renewcommand{\algorithmicrequire}{\textbf{Input:}}
    \renewcommand{\algorithmicensure}{\textbf{Output:}}
    
    \begin{algorithmic}[1]
        \REQUIRE $\boldsymbol{P}$    
        \ENSURE  $\boldsymbol{P}^{'}$   
        
        \STATE  Select a reference point $\boldsymbol{p}_{r}$ from $\boldsymbol{P}$.
        \STATE  Estimate the normal vector of $\boldsymbol{p}_{r}$ according to Formula \eqref{F1} and Formula \eqref{F2}.
        \STATE  Search top $C$ nearest neighbor points of $\boldsymbol{p}_{r}$ in $\boldsymbol{P}$ that adds the weights according to Formula \eqref{F6}.
        \STATE  Calculate local density $\rho_r$ and the total stretched height $D_r$ according to Formula \eqref{F3} and Formula \eqref{F4}.
        \STATE  Determine the stretched direction $dir$.
        \STATE  Stretch to obtain $\boldsymbol{P}^{'}$ according to Formula \eqref{F5}.
        
        \RETURN $\boldsymbol{P}^{'}$
    \end{algorithmic}
\end{algorithm}

\subsubsection{Backbone Adaptation}
After generating synthetic anomalies, we adapt the pre-trained feature extractor to increase its suitability to our targeted industrial data. Specifically, we employ PointMAE~\cite{pointmae} as the extractor $\boldsymbol{f}$.
Anomaly data $\boldsymbol{P}^{'}$ and the corresponding point-wise mask $\boldsymbol{y}$ are obtained from the previous section.

Here, the feature extractor is adapted through supervised learning on the synthetic anomalies. Specifically, we append a segmentation head $\boldsymbol{g}$ after the extractor $\boldsymbol{f}$ to directly predict the anomaly probability $\boldsymbol{e}$ for individual points:
\begin{equation}
\boldsymbol{e} = \boldsymbol{g}(\boldsymbol{f}(\boldsymbol{P}^{'})) \label{F7}
\end{equation}

During the training process, IoU~\cite{iou_loss} and Focal~\cite{Focal_loss} losses are utilized for adapting the backbone:
\begin{equation}
\mathcal{L} = IoU(\boldsymbol{e}, \boldsymbol{y}) + Focal(\boldsymbol{e}, \boldsymbol{y}) \label{F8}
\end{equation}

\subsection{Stage-II Global-Local Memory Bank Construction}

Feature confusion is caused by the local feature similarity of different class data. Considering global features are more descriptive and local features are more sensitive to anomalies, we adopt a global to local strategy to achieve accurate anomaly detection while avoiding feature confusion. In this section, the global memory bank $\boldsymbol{\mathcal{M}}_{G}$ is obtained by clustering the global features of the training data, and then the local memory banks $\boldsymbol{\mathcal{M}}_{i}$ are separately built for each cluster center in $\boldsymbol{\mathcal{M}}_{G}$, as shown in Fig.~\ref{fig:method_overview} b).

\subsubsection{Local and Global Feature Extraction}

Assuming the output from the $i$-th transformer layer of the feature extractor is $\boldsymbol{F}^i$, and denoting $\boldsymbol{F}_l$ as the local features of the input data:

\begin{equation}
    \boldsymbol{F}_l = cat([\boldsymbol{F}^1,\boldsymbol{F}^2,\cdots,\boldsymbol{F}^m]) \label{F9}
\end{equation}
where $m$ is the number of feature layers utilized.

Define $\boldsymbol{F}_g$ is the global features of input data and obtained through average pooling $\boldsymbol{F}_l$:

\begin{equation}
    \boldsymbol{F}_g = AvePooling(\boldsymbol{F}_l) \label{F10}
\end{equation}

$\boldsymbol{F}_l$ describes the local geometric features, while $\boldsymbol{F}_g$ describes the overall shape attributes.

\subsubsection{Global Memory Bank Establishment}
As indicated in the Fig.~\ref{fig:method_overview} b), $K$ centers $\boldsymbol{F}_{ci}, 1 \leq i \leq K$ are obtained by clustering the global features of all training data:

\begin{equation}
    {\boldsymbol{F}_{c1},\boldsymbol{F}_{c2},\cdots,\boldsymbol{F}_{cK}} = Cluster({\boldsymbol{F}_{g1},\boldsymbol{F}_{g2},\cdots,\boldsymbol{F}_{gN}}) \label{F11}
\end{equation}

\noindent where $N$ is the total number of training data, $\boldsymbol{F}_{gj}$ is the global feature of $j$-th training data. The global memory bank $\boldsymbol{\mathcal{M}}_{G}$ is constructed by $K$ cluster centers.

\subsubsection{Local Memory Bank Establishment}
For $j$-th training data, its global features are leveraged to query $\boldsymbol{\mathcal{M}}_{G}$ and determine the index $idx$ of the nearest cluster center:

\begin{equation}
    idx = \mathop{\arg\min}\limits_{i}||\boldsymbol{F}_{ci}-\boldsymbol{F}_{gj}|| \label{F12}
\end{equation}

Then the local feature $\boldsymbol{F}_{lj}$ of the $j$-th training data is delivered into the corresponding $idx$-th local memory bank to establish $\boldsymbol{\mathcal{M}}_{idx}$.

The construction process of  $\boldsymbol{\mathcal{M}}_{idx}$  adopts the coreset strategy:
\begin{equation}
    \boldsymbol{\mathcal{M}}_{idx} = \mathop{\arg\min}\limits_{\boldsymbol{\mathcal{M}}_{idx}\subset\boldsymbol{\mathcal{M}}^{All}_{idx}}  \mathop{\max}\limits_{s\in\boldsymbol{\mathcal{M}}_{idx}^{All} }
    \mathop{\min}\limits_{t\in\boldsymbol{\mathcal{M}}_{idx}}   ||s-t|| \label{F13}
\end{equation}

\noindent where $\boldsymbol{\mathcal{M}}_{idx}^{All}$ is the set of  all local features corresponding to $idx$-th  cluster center.
$\boldsymbol{\mathcal{M}}_{idx}$ is a subset of $\boldsymbol{\mathcal{M}}_{idx}^{All}$, ensuring that the any element in $\boldsymbol{\mathcal{M}}_{idx}$ are farthest from the nearest elements in $\boldsymbol{\mathcal{M}}_{idx}^{All}$. Therefore, for any memory bank $\boldsymbol{\mathcal{M}}_{idx}^{All}$, a coreset $\boldsymbol{\mathcal{M}}_{idx}$ with a specific size can be obtained. The coreset can represent the distribution of original features and reduce unnecessary computational cost.

\subsection{Stage-III Anomaly Detection}

In the test process, test data are queried in $\boldsymbol{\mathcal{M}}_{G}$ and $\boldsymbol{\mathcal{M}}_{idx}$ by their global features and local features to get the object-wise and point-wise anomaly scores, as indicated by the red arrow in the Fig.~\ref{fig:method_overview} c).

\subsubsection{Global Cluster Query}

For any input point cloud $\boldsymbol{P}_{test}$, global and local features are extracted by $\boldsymbol{f}$:

\begin{equation}
\left\{
\begin{aligned}
\boldsymbol{F}_{l}^{test}  &= \boldsymbol{f}(\boldsymbol{P}_{test})   \\
\boldsymbol{F}_{g}^{test} &= AvePooling(\boldsymbol{F}_{l}^{test})\\
\end{aligned}
\right.\label{F16}
\end{equation}

The corresponding $idx$ can be obtained by querying $\boldsymbol{\mathcal{M}}_{G}$ using $\boldsymbol{F}_{g}$:

\begin{equation}
    idx = \mathop{\arg\min}\limits_{i}||\boldsymbol{F}_{ci}-\boldsymbol{F}_{g}^{test}|| \label{F17}
\end{equation}

\subsubsection{Local Anomaly Detection}

$\boldsymbol{F}_{l}^{test}$ is leveraged to search in $\boldsymbol{\mathcal{M}}_{idx}$ to calculate the nearest feature $\boldsymbol{F}^{*}$:

\begin{equation}
    \boldsymbol{F}^{*} = \mathop{\arg\min}\limits_{s\in \boldsymbol{\mathcal{M}}_{idx}}||s-\boldsymbol{F}_{l}^{test}|| \label{F18}
\end{equation}

The point-wise anomaly score $\boldsymbol{A}$ can be calculated using the distance between $\boldsymbol{F}^{*}$ and $\boldsymbol{F}_{l}^{test}$:

\begin{equation}
    \boldsymbol{A} = ||\boldsymbol{F}_{l}^{test}-\boldsymbol{F}^{*}|| \label{F19}
\end{equation}

The maximum point-wise anomaly score is selected as the object-wise anomaly score $\xi$:
\begin{equation}
    \xi = max(\boldsymbol{A}) \label{F20}
\end{equation}

\section{Experiment}\label{sec:exp}
\subsection{Experimental Settings}
\subsubsection{Dataset}

The experiments are conducted on MVTec 3D-AD~\cite{mvtec3d} and Real3D-AD~\cite{read3d}. MVTec 3D-AD and Real3D-AD contain point clouds scanned by industrial 3D sensors from 10 and 12 different classes, respectively. Each object class consists of normal training data and test data including normal samples and abnormal samples. Accurate ground truth annotations including point-wise mask and object-wise label are provided for individual anomaly test data. In particular, the training and testing data of MVTec 3D-AD are both 2.5D data. The training data of Real3D-AD are real 3D data, while the testing data have only one view and are converted into 2.5D data in the following experiments.

\subsubsection{Implementation Details}
We employ PointMAE\footnote{https://github.com/Pang-Yatian/Point-MAE} with 12 transformer layers pre-trained on ShapeNet~\cite{Shapenet} as our default point cloud feature extractor. Anomaly data are synthesized from normal data in MVTec 3D-AD, with $dx, dy, dz \sim U(0.8,1.2)$ and $C\sim U(n/100,n/50)$. The feature extractor is performed self-supervised learning to adapt parameters in the synthetic anomaly data, totaling 4000 iterations. A single-layer fully connected layer is employed as the segmentation head in adaptation. Background points from MVTec 3D-AD are filtered out, following the setting of BTF~\cite{BTF}. A unified evaluation is used for all class data, without any additional hyperparameters between different classes. All experiments are conducted on a GPU RTX A6000.

\begin{table*}[]
\setlength{\tabcolsep}{8pt}
\centering
\caption{\textbf{Quantitative Results on MVTec 3D-AD Dataset~\cite{mvtec3d}}. The results are presented in O-ROC\%/P-ROC\%/P-PRO\%. The Best Is In \textbf{Bold}, And The Second Best In \underline{Underlined}.}
\setlength\tabcolsep{9.0pt}
\resizebox{1\linewidth}{!}{
\begin{tabular}{c|cccc|
>{\columncolor{blue!8}}c }
\toprule[1.5pt]
Method~$\rightarrow$   & BTF~\cite{BTF}  & M3DM~\cite{M3DM} & Shape-Guided~\cite{shape} & CPMF~\cite{cpmf} & \textbf{\ourmethod{}}         \\
Category~$\downarrow$ & CVPRW'2023 & CVPR'2023 & ICML'2023 &
 PR'2024 & \textbf{Ours} \\
\midrule
Bagel        & 71.0/97.3/93.3       & 78.7/96.2/88.5   & \underline{92.4}/\textbf{99.0}/\textbf{96.9} & 92.1/95.4/88.6 & \textbf{96.2}/\underline{98.9}/\underline{96.7} \\
Cable Gland & 50.3/96.2/87.5        & 62.1/94.5/81.2   & 70.7/95.7/85.1 & \textbf{91.9}/\textbf{97.7}/\textbf{91.9}  & \underline{79.7}/\textbf{97.7}/\underline{90.7} \\
Carrot       & 78.3/99.6/97.7      & 60.5/98.6/94.5 & 97.2/\textbf{99.8}/\textbf{98.1}  & \textbf{99.0}/99.5/97.4 & \textbf{99.0}/\textbf{99.8}/\textbf{98.1} \\
Cookie       & 72.7/90.7/82.9      & \textbf{99.7}/93.0/88.4 & 98.3/\underline{93.4}/\underline{89.5}  & 97.8/89.9/80.5  & \underline{99.2}/\textbf{93.5}/\textbf{89.7} \\
Dowel        & 90.8/94.9/85.8      & 79.5/94.3/83.7   & \textbf{96.4}/\textbf{95.9}/\textbf{89.7} & 92.8/95.1/86.3  & \underline{95.7}/\underline{95.8}/\underline{88.3} \\
Foam         & 53.2/93.3/74.8     & 74.9/92.5/73.7   & 72.4/93.0/77.7 & \underline{76.4}/\textbf{94.2}/\textbf{78.2}  & \textbf{84.3}/\underline{94.0}/\underline{77.8} \\
Peach        & 57.6/98.7/95.4      & 66.6/95.5/81.7   & \textbf{94.3}/\underline{99.3}/\underline{97.4} & 92.3/98.7/94.8 & \underline{93.6}/\textbf{99.8}/\textbf{98.1} \\
Potato       & 64.2/99.9/98.3     & 49.6/98.2/93.5   & 90.5/\textbf{99.9}/\textbf{98.3} & \underline{98.2}/99.7/98.0  & \textbf{98.8}/\textbf{99.9}/\textbf{98.3} \\
Rope         & 93.2/98.9/92.8      & 91.6/99.1/92.8   & \underline{97.6}/\underline{99.3}/\underline{95.0} & 94.9/98.7/92.4 & \textbf{98.6}/\textbf{99.6}/\textbf{95.7} \\
Tire         & 50.2/98.8/95.5      & 60.5/98.5/94.5   & \underline{88.6}/\underline{99.5}/\underline{96.8} & 82.2/98.9/96.0 & \textbf{99.1}/\textbf{99.7}/\textbf{97.8} \\ \midrule
Mean         & 68.2/96.8/90.4      & 72.4/96.0/87.2   & 89.8/\underline{97.5}/\underline{92.5}  & \underline{91.8}/96.8/90.4 & \textbf{94.4}/\textbf{97.9}/\textbf{93.1} \\ 
\bottomrule[1.5pt]
\end{tabular}
}
\label{table:mvtec}
\end{table*}

\begin{figure*}[t!]
\centering\includegraphics[scale=1.0]{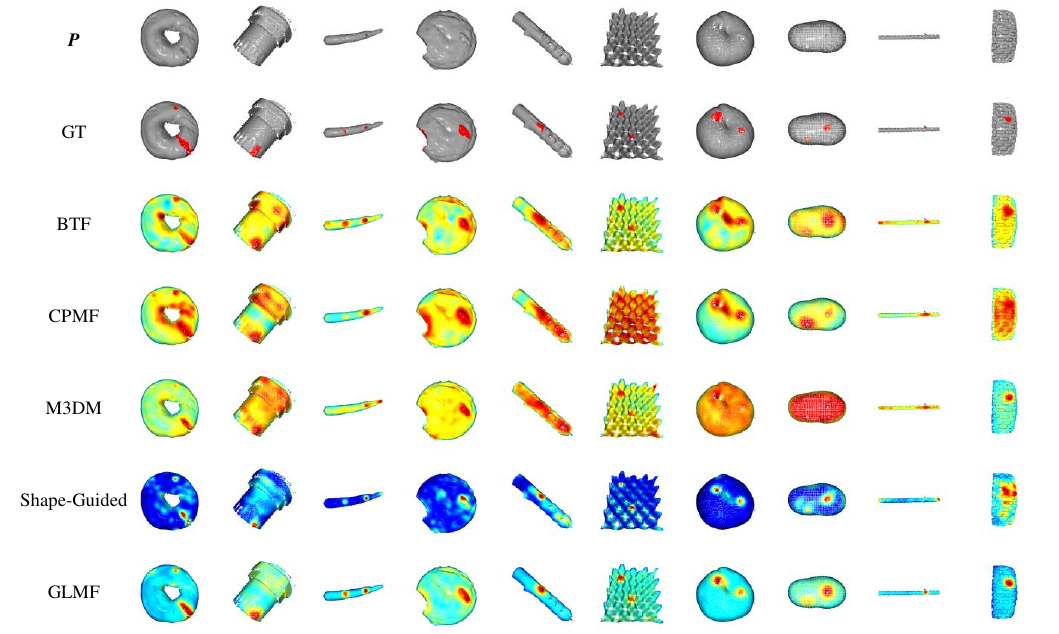}
\caption{\textbf{Visualization of prediction results in MVTec 3D-AD dataset using the proposed method and other methods.} The first row is the original point clouds, while the second row is the ground truth. Subsequent rows depict various methods.}
\label{fig:mvtec}
\end{figure*}

\subsubsection{Evaluation Metrics}

Several key metrics are used to evaluate the performance of anomaly detection, namely the Area Under the Receiver Operating Characteristic curve (AUROC) and Area Under Per Region Overlap (AUPRO). AUROC is utilized for both object-wise and point-wise anomaly detection evaluation, denoted as O-ROC and P-ROC, respectively. AUPRO is exclusively employed for point-wise anomaly detection evaluation and is evaluated in MVTec 3D-AD and Real3D-AD by projecting the point clouds to 2D images, denoted as P-PRO.

\subsection{Comparison Studies}

To comprehensively investigate the efficacy of the proposed multi-class point cloud anomaly detection method, this paper systematically assesses various methods, including BTF~\cite{BTF}, CPMF~\cite{cpmf}, M3DM~\cite{M3DM} and Shape-Guided~\cite{shape}. The aforementioned methods are previous state-of-the-art (SOTA) methods for point cloud anomaly detection and are evaluated on MVTec 3D-AD. It is worth noting that existing 3D anomaly detection methods typically evaluate their detection performance by projecting 3D data into 2D images. While this solution is applicable for 2.5D datasets such as MVTec 3D-AD, it cannot be employed for real 3D datasets such as the training data of Real3D-AD. Hence, only BTF, M3DM and \ourmethod{} are evaluated on Real3D-AD due to the real 3D data format. In addition, 2D pooling in BTF and M3DM has been replaced by point cloud pooling that performs average pooling on neighboring points. We evaluate the multi-class anomaly detection performance separately on MVTec 3D-AD and Real3D-AD, respectively. To provide a more general evaluation of method performance, we mix 22 classes from the two datasets and assess the performance of these methods.

\subsubsection{Comparison results on MVTec 3D-AD}

Quantitative results on MVTec 3D-AD are presented in Table~\ref{table:mvtec}. The number of clusters is selected as 10 in the proposed method \ourmethod{}. \ourmethod{} achieves the best performance in six and seven classes for object-wise and point-wise anomaly detection, respectively, and attains the second best performance in the remaining classes, surpassing +2.6\% object-wise AUROC and +0.6\% point-wise AUPRO compared with the second-placed method in mean performance. The visualized results are displayed in Fig.~\ref{fig:mvtec}. \ourmethod{} accurately detects anomalies with few misses or false positives. BTF utilizes manually designed FPFH features that contain insufficient point information, thus resulting in poor anomaly detection performance. The other three methods all utilize pre-trained models to extract features. However, due to the feature confusion between classes, many false positives occur in certain areas. Through adaptation with synthetic anomalies, the feature extractor of \ourmethod{} is encouraged to focus on local geometric changes in point clouds. The global-local memory bank also mitigates feature confusion among different classes, yielding optimal performance.

The processing speeds of all methods are calculated by accounting for the pre-processing and detection times for 1197 test samples from the MVTec 3D-AD, as detailed in Table~\ref{table:speed}. CPMF reaches 6.56s/per because multi-view rendering takes a long time. The feature extraction of M3DM is relatively time-consuming, with an average speed of 3.69s/per. Shape-Guided and BTF demonstrate better efficiency, achieving processing times of 1.49s/per and 1.65s/per, respectively. 
Benefiting from the lightweight feature extractor, which actually only includes two transformer layers, \ourmethod{} does not require any additional processing of the data, achieving an excellent inference speed of 1.38s/per. In summary, \ourmethod{} has achieved state-of-the-art performance and efficiency compared with other methods.

\begin{table}[]
\setlength{\tabcolsep}{3pt}
\centering
\caption{Speed of Methods on MVTec 3D-AD Dataset.}
\resizebox{\linewidth}{!}{
\begin{tabular}{@{}cccccc@{}}
\toprule[1.5pt]
Statistic~$\rightarrow$       & Pre-process  & Detect & Total Time & Speed  \\ 
Method~$\downarrow$ & (s) & (s)  &  (s) & (s/per)\\

\midrule
BTF~\cite{BTF}      & -      & 1976         & 1976        & 1.65      \\ 
M3DM~\cite{M3DM}   & -         & 4419        & 4419        & 3.69     \\
Shape-Guided~\cite{shape} & 290       & 1494       & 1784      & 1.49       \\
CPMF~\cite{cpmf} & 4938      & 2911        & 7849        & 6.56      \\

\rowcolor{blue!8} {\ourmethod{}}            &{-}          & {1657}            & {1657}      & {\textbf{1.38}}  \\ 
\bottomrule[1.5pt]
\end{tabular}
}
\label{table:speed}
\end{table}

\subsubsection{Comparison results on Real3D-AD}

\begin{figure}[t!]
\centering\includegraphics[width=\linewidth]{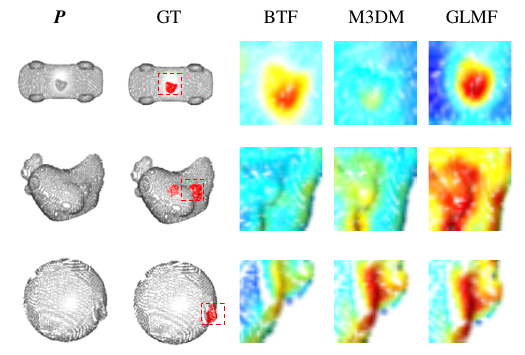}
\caption{\textbf{Visualization of prediction results in Real3D-AD dataset using the proposed method and other methods.} The first column is the original point clouds, while the second column is the ground truth. Subsequent columns depict various methods.}
\label{fig:real}
\end{figure}

Quantitative results on Real3D-AD are presented in Table~\ref{table:real}. The number of clusters is selected as 3 in \ourmethod{}. \ourmethod{} achieves 53.2\% object-wise AUROC, 89.8\% point-wise AUROC and 67.0\% point-wise AUPRO, outperforming existing methods in nearly all classes. Given that the training data comprise complete point clouds scanned from multiple perspectives, while the test data are derived from a single perspective, the contours of the test data are easily misidentified as anomalies. Therefore, the gap between data distributions leads to relatively poor performance of current methods. Benefiting from the adaptation on synthetic anomaly data and the global-local memory bank, \ourmethod{} still achieves strong performance. Selected anomaly areas are visualized in Fig.~\ref{fig:real}. BTF and M3DM exhibit varying degrees of false positives, whereas \ourmethod{} more accurately detects defect areas compared to them.

\begin{table}[]
\centering
\caption{\textbf{Quantitative Results on Real3D-ad Dataset~\cite{read3d}}. The results are presented in O-ROC\%/P-ROC\%/P-PRO\%. The Best Is In \textbf{Bold}, And The Second Best In \underline{Underlined}.}
\resizebox{\linewidth}{!}{
\begin{tabular}{c|cc|
>{\columncolor{blue!8}}c }
\toprule[1.5pt]

Method~$\rightarrow$   & BTF~\cite{BTF}  & M3DM~\cite{M3DM} &  \textbf{\ourmethod{}}         \\
Category~$\downarrow$ & CVPRW'2023 & CVPR'2023 & \textbf{Ours} \\ \midrule

Airplane   & \underline{65.3}/83.9/\underline{56.5} & 33.3/\underline{85.7}/56.2 & \textbf{73.3}/\textbf{86.3}/\textbf{62.5} \\
Candybar   & \textbf{61.4}/91.7/72.3 & 37.6/\underline{92.4}/\underline{74.3} & \underline{59.2}/\textbf{94.4}/\textbf{81.4} \\
Car        & \underline{51.7}/88.8/67.5 & 36.3/\textbf{93.3}/\underline{75.3} & \textbf{53.4}/\underline{92.7}/\textbf{77.7} \\
Chicken    & \underline{41.7}/83.2/43.4 & 38.3/\underline{84.8}/\underline{44.4} & \textbf{57.4}/\textbf{84.9}/\textbf{45.2} \\
Diamond    & \textbf{59.3}/81.6/41.9 & 56.8/\textbf{87.5}/\underline{60.7} & \underline{57.8}/\underline{87.2}/\textbf{60.9} \\
Duck       & \textbf{49.8}/75.0/24.3 & \underline{41.9}/\underline{83.7}/\underline{42.7} & 30.2/\textbf{85.7}/\textbf{55.4} \\
Fish       & 31.8/\underline{92.0}/\underline{69.9} & \underline{51.2}/88.8/64.1 & \textbf{69.3}/\textbf{94.3}/\textbf{80.5} \\
Gemstone   & \textbf{51.6}/79.9/34.0 & \underline{49.1}/\textbf{86.1}/\textbf{53.8} & 34.0/\underline{84.7}/\underline{50.0} \\
Seahorse   & \textbf{62.7}/94.3/80.0 & 9.0/\underline{95.5}/\underline{82.0}  & \underline{47.5}/\textbf{96.8}/\textbf{86.9} \\
Shell      & \underline{36.9}/74.1/25.1 & \textbf{63.1}/\underline{82.7}/\underline{42.6} & 25.1/\textbf{83.6}/\textbf{48.0} \\
Starfish   & 45.9/91.7/74.6 & \underline{49.0}/\textbf{92.9}/\textbf{76.6} & \textbf{56.0}/\underline{92.8}/\underline{75.9} \\
Toffees    & 44.2/76.2/27.7 & \underline{50.2}/\underline{88.3}/\underline{61.1} & \textbf{75.5}/\textbf{93.9}/\textbf{79.3} \\ \midrule
Mean       & \underline{50.2}/84.4/51.4 & 43.0/\underline{88.5}/\underline{61.2} & \textbf{53.2}/\textbf{89.8}/\textbf{67.0} \\ \bottomrule[1.5pt]
\end{tabular}
}
\label{table:real}
\end{table}

\subsubsection{Comparison results on Mix dataset}

To evaluate the performance of the methods in more general scenarios, we mix the MVTec 3D-AD and Real3D-AD to create a new dataset with a total of 22 classes. The quantitative experimental results are presented in Table~\ref{table:mix}. The number of clusters is selected as 13 in \ourmethod{}. With more classes, the feature confusion problem becomes more severe. The features extracted by BTF and M3DM struggle to effectively distinguish between normal and abnormal instances, resulting in very poor object-wise performance. Due to the similarity between normal and abnormal features, BTF and M3DM also encounter significant difficulties in point-wise anomaly detection.
Due to the effective suppression of feature confusion through global-local feature matching, \ourmethod{} has achieved 72.9\% object-wise AUROC, 93.1\% point-wise AUPRO, and 77.9\% point-wise AUPRO, significantly surpassing BTF and M3DM.

\begin{table}[t]
\centering
\caption{\textbf{Quantitative Results on Mix Dataset.} The results are presented in O-ROC\%/P-ROC\%/P-PRO\%. The Best Is In \textbf{Bold}, And The Second Best In \underline{Underlined}.}
\resizebox{\linewidth}{!}{
\begin{tabular}{c|cc|
>{\columncolor{blue!8}}c }
\toprule[1.5pt]
Method~$\rightarrow$   & BTF~\cite{BTF}  & M3DM~\cite{M3DM} &  \textbf{\ourmethod{}}         \\
Category~$\downarrow$ & CVPRW'2023 & CVPR'2023 & \textbf{Ours} \\ \midrule

{MVTec 3D-AD} &  67.0/\underline{96.8}/\underline{90.4}   & \underline{72.2}/96.1/87.3 & \textbf{93.7}/\textbf{97.8}/\textbf{92.8}  \\ 
                 
{Real3D-AD} &  \underline{53.3}/84.6/51.7   & 42.9/\underline{88.5}/\underline{61.2} & \textbf{66.7}/\textbf{89.2}/\textbf{65.5} \\ 

{Mean}& \underline{60.2}/90.7/71.1   & 57.6/\underline{92.3}/\underline{74.3} & \textbf{80.2}/\textbf{93.5}/\textbf{79.2}   \\
\bottomrule[1.5pt]
\end{tabular}
}
\label{table:mix}
\end{table}

\begin{figure*}[t!]
\centering\includegraphics[width=\linewidth]{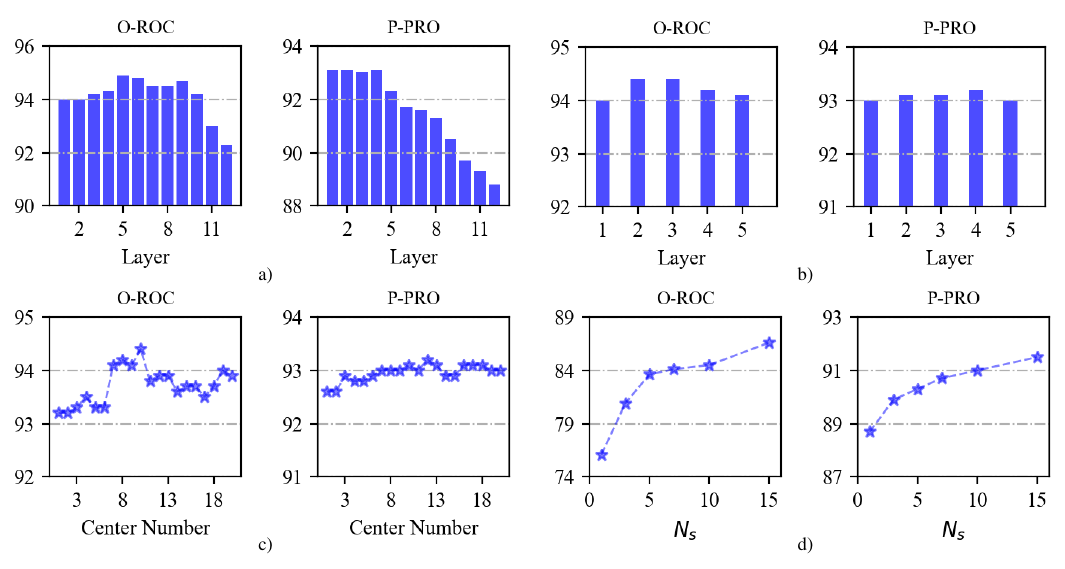}
\caption{
a) The performance using outputs from different transformer layers. The horizontal axis represents the transformer layer selected. b) The performance using composite outputs from different transformer layers. Number $i$ in horizontal axis represents the first $i$ transformer layers are selected. c) The performance using different center number. The horizontal axis represents the center number used. d) The performance on MVTec 3D-AD under different few-shot settings. The horizontal axis represents the sample number of each class.
}
\label{fig:four}
\end{figure*}

\subsection{Ablation Studies}

In this subsection, we verify the effectiveness of the proposed two technologies: adaptation on synthetic anomaly data
and global-local memory bank (GLMB). Additionally, we explore the impact of various levels of features from the feature extractor on performance, and identify the feature space distribution of test data utilizing the adapted feature extractor or not. Finally, the work mechanism of GLMB is detailed by the visualization of feature clustering and anomaly score distribution. Owing to the limited training data in the Real3D-AD, all experiments in this subsection are conducted on the MVTec 3D-AD. Considering the inflated point-wise AUROC in MVTec 3D-AD, only object-wise AUROC and point-wise AUPRO are used for performance evaluation.

Table~\ref{table:module} shows the quantitative results under different settings. It can be observed that both adaptation and GLMB can improve performance because of better feature discrimination and weaker feature confusion. When both modules are adopted, the performance reaches best, with an improvement of +8.1\% object-wise AUROC and +1.7\% point-wise AUPRO compared with that without modules. Below we deeply analyze the working mechanisms of the proposed modules.

\begin{table}[]
\setlength{\tabcolsep}{15pt}

\centering
\caption{\textbf{Quantitative Results on MVTec 3D-ad Dataset Under Different Setting.} The Red Fonts Represent A Performance Increment Compared With The Setting Without Any Proposed Modules.(\%)}
\setlength\tabcolsep{18.0pt}
\resizebox{\linewidth}{!}{
\begin{tabular}{@{}cc|cc@{}}
\toprule[1.5pt]
{Adaptation}  &      {GLMB}    &    {O-ROC}  & {P-PRO}         \\ \midrule
{\color{black}\ding{55}}       & {\color{black}\ding{55}}     &    86.1 & 91.4            \\
{\color{black}\ding{51}} & {\color{black}\ding{55}}      &    93.2 \scriptsize{\textcolor{red}{(+7.1)}} & 92.6  \scriptsize{\textcolor{red}{(+1.2)}}  \\
{\color{black}\ding{55}} & {\color{black}\ding{51}}      &    87.6 \scriptsize{\textcolor{red}{(+1.5)}} & 92.0 \scriptsize{\textcolor{red}{(+0.6)}}    \\
{\color{black}\ding{51}} & {\color{black}\ding{51}}  &     94.4 \scriptsize{\textcolor{red}{(+8.3)}} & 93.1 \scriptsize{\textcolor{red}{(+1.7)}} \\    
\bottomrule[1.5pt]
\end{tabular}
}
\label{table:module}
\end{table}

\subsubsection{Influence of feature extractor layer}

Features at different levels can be extracted by various transformer layers of the feature extractor. To verify the discriminative ability of different level features, we utilize the output from each transformer layer of the feature extractor to establish memory banks and perform anomaly detection, as illustrated in Fig.~\ref{fig:four} a). With the use of deeper layers, object-wise performance initially improves, then stabilizes at a higher level before finally decreasing. Point-wise performance is initially high but subsequently decreases. The degradation in performance is attributed to the high-level features of the feature extractor containing biases toward pre-trained data, potentially harming downstream tasks like anomaly detection. Consequently, shallow features containing rich local geometric information are more beneficial for anomaly detection. This insight leads us to employ only the first $i$ layers for this task. The results are presented in Fig.~\ref{fig:four} b). It is observed that using only the first two layers of features yields better performance both object-wise and point-wise. Moreover, during testing, only two transformer layers need to be computed, significantly improving efficiency.

\subsubsection{Influence of Adaptation}
\begin{figure}[t]
\centering\includegraphics[width=\linewidth]{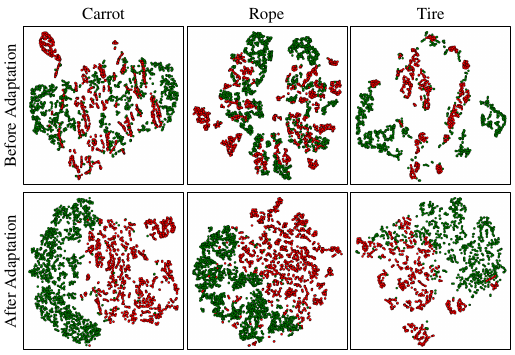}
\caption{\textbf{Visualization of feature distribution from multi-class point clouds utilizing pre-trained PointMAE (Before Adaptation) and adapted feature extractor (After Adaptation).} The red points represent anomalous features while the green points represent normal features.}
\vspace{-3mm}
\label{fig:single_cluster}
\end{figure}

To more intuitively explore the influence of adaptation on performance improvement, we extract the features from the multi-class test point clouds leveraging pre-trained PointMAE and adapted feature extractor respectively, and visualize normal and anomalous features using t-SNE method. As shown in Fig.~\ref{fig:single_cluster}, there is clearly a large overlap between the normal and anomalous features extracted by PointMAE, indicating that PointMAE cannot confidently distinguish between normal and anomalous points. On the contrary, the normal and anomalous features extracted by the adapted feature extractor are easily separated by a clear boundary, not only in the simple Carrot class, but even in the challenging Rope class and Tire class, meaning it can accurately understand and capture anomaly areas and obtain highly discriminative features. Furthermore, the excellent feature extractor also reduces the probability of feature confusion in multi-class anomaly detection task.

\begin{figure}[h!]
\centering\includegraphics[width=0.95\linewidth]{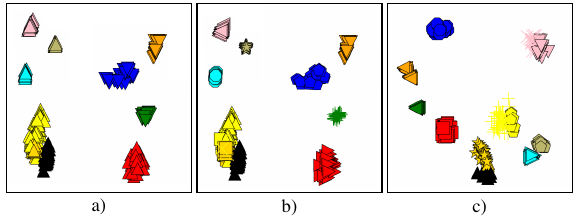}
\caption{\textbf{Visualization of ten classes data from MVTec 3D-AD dataset and cluster centers.} Ten colors represent different classes data, and different shapes represent different cluster centers. a) $K$=2 and using the adapted feature extractor. b) $K$=10 and using the adapted feature extractor. c) $K$=10 and using the pre-trained PointMAE.}
\vspace{-3mm}
\label{fig:cluster}
\end{figure}

\subsubsection{Influence of GLMB}

\begin{figure}[t]
\centering\includegraphics[width=\linewidth]{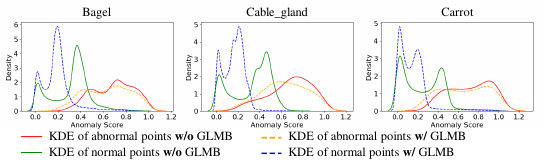}
\caption{\textbf{Anomaly Score Distribution w/o and w/ the proposed GLMB.} Anomaly scores are min-max normalized for better visualizations. }
\vspace{-3mm}
\label{fig:kde}
\end{figure}

GLMB utilizes a larger receptive field of global features to separate point clouds that are prone to feature confusion. Therefore, the appropriate number of clustering centers can better leverage the advantages of GLMB to reduce the probability of feature confusion, without affecting the performance of local feature anomaly detection.
The anomaly detection results using different numbers of cluster centers are illustrated in Fig.~\ref{fig:four} c). As the number of cluster centers increases, the performance of the object-wise detection firstly remains unchanged, and then gradually increases and fluctuates at a higher level. The performance of the point-wise detection is also maintained at first, and then increases and stabilizes at a higher level. When the number of cluster centers is small, as shown in Fig.~\ref{fig:cluster} a), only some classes are individually clustered, while others still exhibit significant feature confusion. When there are many cluster centers, as shown in Fig.~\ref{fig:cluster} b), seven classes have been correctly clustered. The reason why the other three classes' data are not well clustered is that they are very similar and the types of anomalies are also similar. Therefore, in fact, the data of these three classes are clustered based on more detailed criteria than class alone, which makes it easier to distinguish anomalies.
The robustness of GLMB is evident as performance remains high when the number of cluster centers is close to optimal.
The results from the feature extractor without adaptation are disordered and unreliable, as depicted in Fig.~\ref{fig:cluster} c), underscoring the crucial need for effective global features in GLMB.
Additionally, we employ Kernel Density Estimation (KDE) to visualize the anomaly scores of the test data, as shown in Fig.~\ref{fig:kde}. Ideally, the anomaly scores for anomalous points should be uniformly lower than those for normal points, with the overlap between the two curves signifying the detection error. Utilizing GLMB, the overlap area significantly decreased, indicating that the features that were previously difficult to distinguish due to feature confusion have become more distinguishable. GLMB has improved detection performance in almost all classes.

\subsection{Few-shot Studies}

Few-shot anomaly detection means training with limited data, and the risk of feature confusion can drastically impair the performance of multi-class detection. We evaluate the performance of \ourmethod{} in a few-shot multi-class task in this section. The number of training data for each class $N_s$ is set to 1, 3, 5, 7, 10 and 15, while all data are mixed for training. The test results are shown in Fig.~\ref{fig:four} d). \ourmethod{} also exhibits strong performance with performance gradually improving as the amount of data increases. Notably, \ourmethod{} even has already exceeded 90\% at P-PRO with ten training samples.

\subsection{Actual Inspection on Industry Parts}

\begin{figure}[t!]
\centering\includegraphics[width=\linewidth]{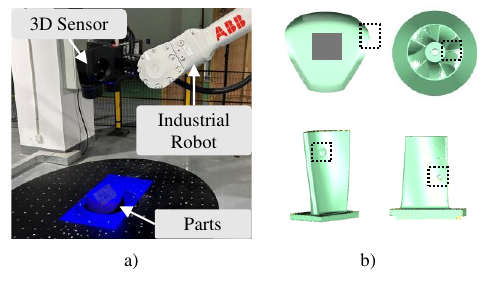}
\caption{a) The point cloud collection process of actual industry parts. b) Some abnormal samples from four object classes.}
\label{fig:scan}
\end{figure}

To further evaluate the actual performance of the \ourmethod{}, inspection experiments are conducted on actual industry parts, encompassing four classes: Blade\_1 and Blade\_2, Impeller and Plastic\_part. The scanning process is shown in Fig.~\ref{fig:scan} a), with the brand of the plastic parts masked. An industry parts point cloud anomaly detection dataset is scanned by an industrial robot equipped with 3D sensors. In the dataset, Blade\_1, Blade\_2, and Impeller have 30 normal samples and 30 abnormal samples respectively, while the Plastic\_part has 50 normal samples and 50 abnormal samples. Some abnormal data are depicted in Fig.~\ref{fig:scan} b).

\begin{figure}[t!]
\centering\includegraphics[width=\linewidth]{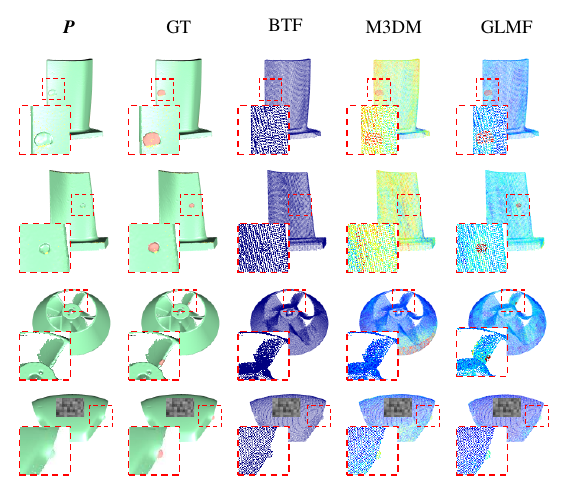}
\caption{\textbf{Visualization of prediction results in actual industry parts dataset using the proposed method and other methods.} The first column is the original point clouds, while the second column is the ground truth. Subsequent columns depict various methods.}
\vspace{-3mm}
\label{fig:actual}
\end{figure}

Quantitative results are presented in Table~\ref{table:actual}. BTF has almost failed in all classes because industrial parts typically have surfaces with smaller curvature, which makes FPFH features close to any local points. In the Blade\_1 class, M3DM only performs well on object-wise performance but poor on point-wise performance. In the other classes, due to feature confusion, normal and abnormal data in the test data are mistakenly identified as the abnormal and normal of other classes, resulting in AUROC scores below 50\%. However, the performance of \ourmethod{} reaches 82.4\% object-wise AUROC and 90.2\% point-wise AUROC, significantly higher than the other two methods +25.5\% and +37.5\%. The visualization results are displayed in Fig.~\ref{fig:actual}, and the anomaly areas are highlighted with red boxes. It is evident that \ourmethod{} effectively identifies anomalies, achieving the required standard for actual industrial anomaly detection.

\begin{table}[]
\centering
\caption{\textbf{Quantitative Results on Actual Industry Parts Dataset.} The results are presented in O-ROC\%/P-ROC\%. The Best Is In \textbf{Bold}.}
\resizebox{\linewidth}{!}{
\begin{tabular}{c|cc|
>{\columncolor{blue!8}}c }
\toprule[1.5pt]
Method~$\rightarrow$   & BTF~\cite{BTF}  & M3DM~\cite{M3DM} &  \textbf{\ourmethod{}}         \\
Category~$\downarrow$ & CVPRW'2023 & CVPR'2023 & \textbf{Ours} \\ \midrule

Blade\_1      & 56.7/49.7   & 98.9/66.4 & \textbf{100.0}/\textbf{98.2} \\
Blade\_2      & 51.7/49.9   & 44.7/50.3 & \textbf{73.3}/\textbf{92.2}  \\
Impeller      & 58.3/50.4 & 42.0/47.2  & \textbf{58.8}/\textbf{93.9}  \\
Plastic\_part & 52.0/50.0 & 42.1/47.0 & \textbf{97.6}/\textbf{76.3}  \\ \midrule
Mean          & 54.7/50.0 & 56.9/52.7 & \textbf{82.4}/\textbf{90.2}  \\ \bottomrule[1.5pt]
\end{tabular}
}
\label{table:actual}
\end{table}

\subsection{Discussion}
The above experiments demonstrate that our method \ourmethod{} is effective on both public and actual datasets, and the ablation experiments also explore in detail the effectiveness and superiority of the proposed technique adaptation and GLMB. Methods like M3DM and Shape-guided utilize multi-level features for anomaly detection, but in fact, shallow features are more effective for products or industrial parts with small curvatures. Non-discriminative high-level features can compromise detection accuracy. However, leveraging only local features may encounter the feature confusion due to their small receptive field. Therefore, to solve this seemingly contradictory problem, we aggregate local features into global features with large receptive fields using clustering to solve feature confusion, without affecting the local features for accurate anomaly detection. 
The primary distinction and innovation of \ourmethod{} relative to other state-of-the-art methods lies in its targeted proposal and resolution of feature confusion.

Despite the superior performance of our method in the experimental evaluations, a notable disparity remains between the performance of point cloud feature extractors and that of image feature extractors, making point cloud anomaly detection methods unable to utilize the recent and advanced frameworks, such as reconstruction and knowledge distillation. Consequently, the future development of more potent self-supervised techniques and the generation of more authentic anomaly data, potentially through generative models like diffusion models, is anticipated to enhance the capabilities of point cloud feature extractors.

\section{Conclusion}\label{sec:conclusion}

This paper proposes \ourmethod{}, a global-local feature matching strategy for multi-class point cloud anomaly detection aimed at addressing feature confusion. Global features are utilized to distinguish among different classes, while local features facilitate both object-wise and point-wise anomaly detection, thereby preventing anomalies of one class from being mistakenly identified as normal in another. Additionally, an anomaly synthesis pipeline is presented to adapt the point cloud feature extractor, yielding more distinct global and local features. Comprehensive experiments demonstrate the effectiveness of \ourmethod{}, which outperforms previous methods on both the MVTec 3D-AD and Real3D-AD. Real-world experiments further confirm the practicality of \ourmethod{}.


\ifCLASSOPTIONcaptionsoff
  \newpage
\fi



%

{\small
\bibliographystyle{unsrt}

\bibliography{ref}
}
\end{document}